\documentclass[letterpaper,twocolumn,10pt]{article}
\usepackage{include/cvpr/cvpr}
\usepackage{times}
\usepackage{epsfig}
\usepackage{graphicx}
\usepackage{amsmath}
\usepackage{amssymb}
\usepackage[breaklinks=true,bookmarks=false,colorlinks=true]{hyperref}
\usepackage{multirow}
\usepackage{inconsolata}

\cvprfinalcopy
\def\cvprPaperID{****}

\newcommand{\tsc}[1]{\textsuperscript{#1}}
\newcommand{\tbf}[1]{\textbf{#1}}
\newcommand{\mbf}[1]{\boldsymbol{#1}}

\newcommand\extrafootertext[1]{%
    \bgroup
    \renewcommand\thefootnote{\fnsymbol{footnote}}%
    \renewcommand\thempfootnote{\fnsymbol{mpfootnote}}%
    \footnotetext[0]{#1}%
    \egroup
}
\title{
Rapid Automated Analysis of Skull Base Tumor Specimens Using\\
Intraoperative Optical Imaging and Artificial Intelligence
}
\author{
Cheng Jiang\tsc{1},
Abhishek Bhattacharya\tsc{1}, Joseph Linzey\tsc{1},
Rushikesh S. Joshi\tsc{1}, Sung Jik Cha\tsc{2},\\
Sudharsan Srinivasan\tsc{1}, Daniel Alber\tsc{3}, Akhil Kondepudi\tsc{1},
Esteban Urias\tsc{1}, Balaji Pandian\tsc{1}, \\
Wajd Al-Holou\tsc{1}, Steve Sullivan\tsc{1}, B. Gregory Thompson\tsc{1},
Jason Heth\tsc{1}, Chris Freudiger\tsc{4}, \\
Siri Khalsa\tsc{1}, Donato Pacione\tsc{5}, John G. Golfinos\tsc{5},
Sandra Camelo-Piragua\tsc{1}, \\
Daniel A. Orringer\tsc{5}, Honglak Lee\tsc{1}, and Todd Hollon\tsc{1}
\\[1em]
\tsc{1}University of Michigan
\tsc{2}Western Michigan University
\tsc{3}Brown University\\
\tsc{4}Invenio Imaging
\tsc{5}New York University
}
\begin{document}
\maketitle
\extrafootertext{Published as journal article in
\href{https://doi.org/10.1227/neu.0000000000001929}{\emph{Neurosurgery 90} (6),
758-767}. Correspondences to \texttt{tocho@med.umich.edu}.}
\begin{abstract}
\textbf{Background:} Accurate specimen analysis of skull base tumors is
essential for providing personalized surgical treatment strategies.
Intraoperative specimen interpretation can be challenging because of the wide
range of skull base pathologies and lack of intraoperative pathology resources.

\textbf{Objective:}  To develop an independent and parallel intraoperative
workflow that can provide rapid and accurate skull base tumor specimen analysis
using label-free optical imaging and artificial intelligence.

\textbf{Methods:} We used a fiber laser–based, label-free, nonconsumptive,
high-resolution microscopy method ($<60$ seconds per $1\times 1\ \text{mm}^2$),
called stimulated Raman histology (SRH), to image a consecutive, multicenter
cohort of patients with skull base tumor. SRH images were then used to train a
convolutional neural network model using 3 representation learning strategies:
cross-entropy, self-supervised contrastive learning, and supervised contrastive
learning. Our trained convolutional neural network models were tested on a
held-out, multicenter SRH data set.

\textbf{Results:} SRH was able to image the diagnostic features of both benign
and malignant skull base tumors. Of the 3 representation learning strategies,
supervised contrastive learning most effectively learned the distinctive and
diagnostic SRH image features for each of the skull base tumor types. In our
multicenter testing set, cross-entropy achieved an overall diagnostic accuracy
of 91.5\%, self-supervised contrastive learning 83.9\%, and supervised
contrastive learning 96.6\%. Our trained model was able to segment tumor-normal
margins and detect regions of microscopic tumor infiltration in meningioma SRH
images.

\textbf{Conclusion:} SRH with trained artificial intelligence models can provide
rapid and accurate intraoperative analysis of skull base tumor specimens to
inform surgical decision-making.

\end{abstract}
\section{Introduction}
Optimal skull base neurosurgery requires personalized surgical treatment
strategies based on clinical, radiographical, and pathological data. Skull base
lesions are diverse and span the full pathology spectrum, including
inflammatory, infectious, and neoplastic diseases. Look-a-like lesions and
uncommon radiographical or clinical features can lead to diagnostic errors
and potentially increase surgical morbidity \cite{altshuler2021imaging,
hollon2016surgical, ierokomos1985primary, kunimatsu2017skull}.
In addition to tumor diagnosis, rapid microscopic assessment of tumor
resection cavities for residual tumor burden could increase gross total
resection and reduce tumor recurrence rates. Residual tumor burden is
the major cause of tumor recurrence in both benign and malignant skull
base tumors \cite{zhai2019nomogram, ueberschaer2021simpson}.
An intraoperative pathology workflow that could provide rapid and
accurate evaluation of skull base surgical specimens has the potential to
guide personalized treatment strategies and improve surgical outcomes.

\begin{figure*}[ht]
\centering
\includegraphics[width=\textwidth]{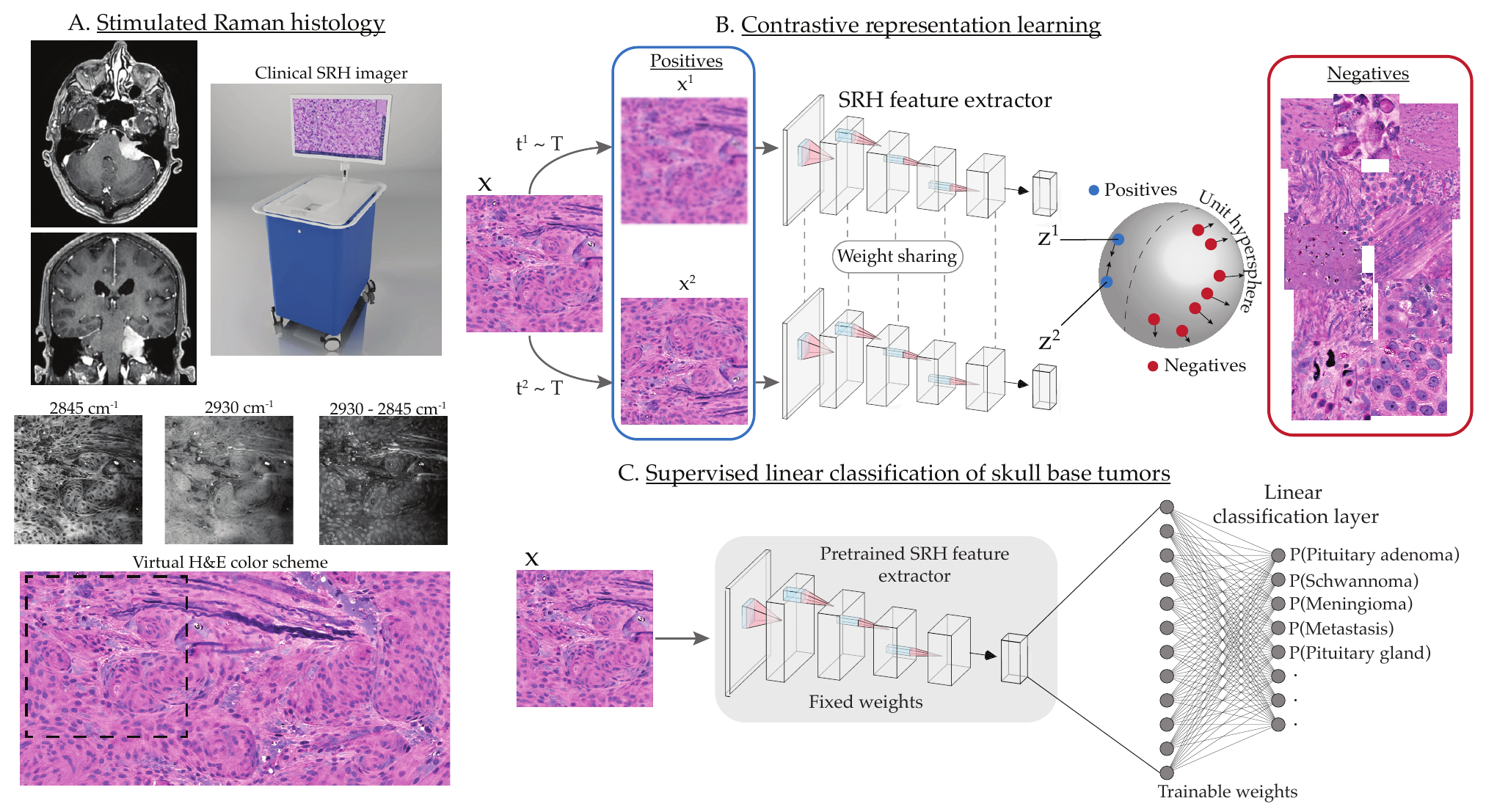}
\caption{Stimulated Raman histology (SRH) and contrastive representation
learning framework. \textbf{A,} Clinical SRH imager used for intraoperative
imaging of fresh brain tumor specimens. The surgical specimen is loaded into a
premade microscope slide. The SRH imager is operated by a single technician
with minimal training through a simple touch-screen interface with prompted
directions. SRH images are acquired by imaging at 2 Raman shifts: 2845 and 2950
cm\tsc{-1}. Lipid-rich regions (eg, myelinated white matter) demonstrate high
SRS signal at 2845 cm\tsc{-1} because of CH2 symmetric stretching in fatty
acids. Cellular regions produce high 2930 cm\tsc{-1} intensity and large
2930:2845 ratios to high protein and nucleic acid content
\cite{orringer2017rapid}. The subtracted image highlights cellularity and
nuclei. A virtual hematoxylin and eosin (H\&E) color scheme is applied to
transform the raw stimulated Raman scattering images into SRH images for
clinical use and pathological review. \textbf{B,} Contrastive representation
learning involves selecting a pair of positive image examples. In the
self-supervised setting, this pair is generated by sampling 2 random
transformations from a set of transformations, $\mathcal{T}$, such as image
blurring ($t^1$) or flipping ($t^1$), and applying the transformations to a
single image, $x$, to get $x^1$ and $x^2$. Both images undergo a feedforward
pass through an SRH feature extractor, which is a convolutional neural network.
$x^1$ and $x^2$ now have normalized vector representations, $z^1$ and $z^2$,
which can then be compared using a similarity metric on the unit hypersphere.
The objective of contrastive learning was to make the similarity metric between
positive examples large and negative examples small. This corresponds to
placing representations of positive pairs near each other and pushing negative
pairs away. In the case of supervised contrastive learning, positive examples
are pairs from the same diagnostic class and negative examples are from all
other classes. \textbf{C,} Finally, after training our SRH feature extractor
using contrastive learning, we train a linear classification layer to provide a
probability distribution over our output classes.}
\label{fig:workflow}
\end{figure*}

Our standard of care for intraoperative assessment of surgical specimens
is based on hematoxylin and eosin (H\&E) staining of processed surgical specimens
and requires interpretation by a board-certified pathologist. Tissue
processing is extensive, requiring transport, staining, sectioning, and
mounting of the specimen. The turnaround times for intraoperative specimen
interpretation (20 - 90 minutes) discourage routine use in skull base
neurosurgery, particularly for tumor margin assessment
\cite{novis1997interinstitutional}. Moreover, the pathology workforce is
contracting, with an overall reduction of 18\% between 2007 and 2017
\cite{robboy2013pathologist, metter2019trends}. In this study,
we propose an alternative workflow for rapid interpretation of surgical
specimens using optical imaging and artificial intelligence (AI).

Stimulated Raman histology (SRH) is a rapid, label-free, high-resolution,
optical imaging method used for intraoperative evaluation of fresh,
unprocessed tissue specimens \cite{freudiger2008label, orringer2017rapid}.
We have previously shown that SRH combined with AI models can
achieve human-level performance for the intraoperative diagnosis of the
most common brain tumor subtypes and recurrent primary brain tumors
\cite{hollon2020near, hollon2021rapid}. Our models detect cytological
and histomorphological features in brain tumors to provide near
real-time diagnoses ($<2$ minutes) without the need for tissue processing
or human interpretation.

In this study, we aim to develop an integrated computer vision system for
rapid intraoperative interpretation of skull base tumors using SRH and AI.
To improve on our previous methods, we applied a new AI training technique,
contrastive representation learning, which boosted our model's ability to
detect diagnostic features in SRH images. We show that this model can
effectively segment tumor-normal margins and detect regions of microscopic
tumor infiltration in grossly normal surgical specimens, allowing for
robust margin delineation in meningioma surgery.


\section{Methods}
\subsection{Study Design}
Study objectives were to (1) determine whether SRH can capture the diagnostic
features of skull base tumors, (2) develop an AI-based computer vision system
that combines clinical SRH and deep neural networks to achieve human-level
performance on the intraoperative classification of skull base tumors, and
(3) demonstrate the feasibility of using our model to detect microscopic tumor
infiltration in meningioma surgery. After Institutional Review Board approval
(HUM00083059), this study began on June 1, 2015. Inclusion criteria were the
following: (1) patients with planned brain tumor resection, including skull
base surgery, at Michigan Medicine (UM) and New York University; (2) subject
or durable power of attorney able to give informed consent; and (3) subjects
in whom there was additional specimen beyond what was needed for routine
clinical diagnosis. We then trained and validated a benchmarked convolutional
neural network (CNN) architecture (ResNets\cite{he2016deep}) on the
classification of fresh surgical specimens imaged with SRH. CNN performance
was then tested using a held-out, multicenter (UM and NYU) prospective testing
SRH data set.


\begin{figure*}[ht]
\centering
\includegraphics[width=\textwidth]{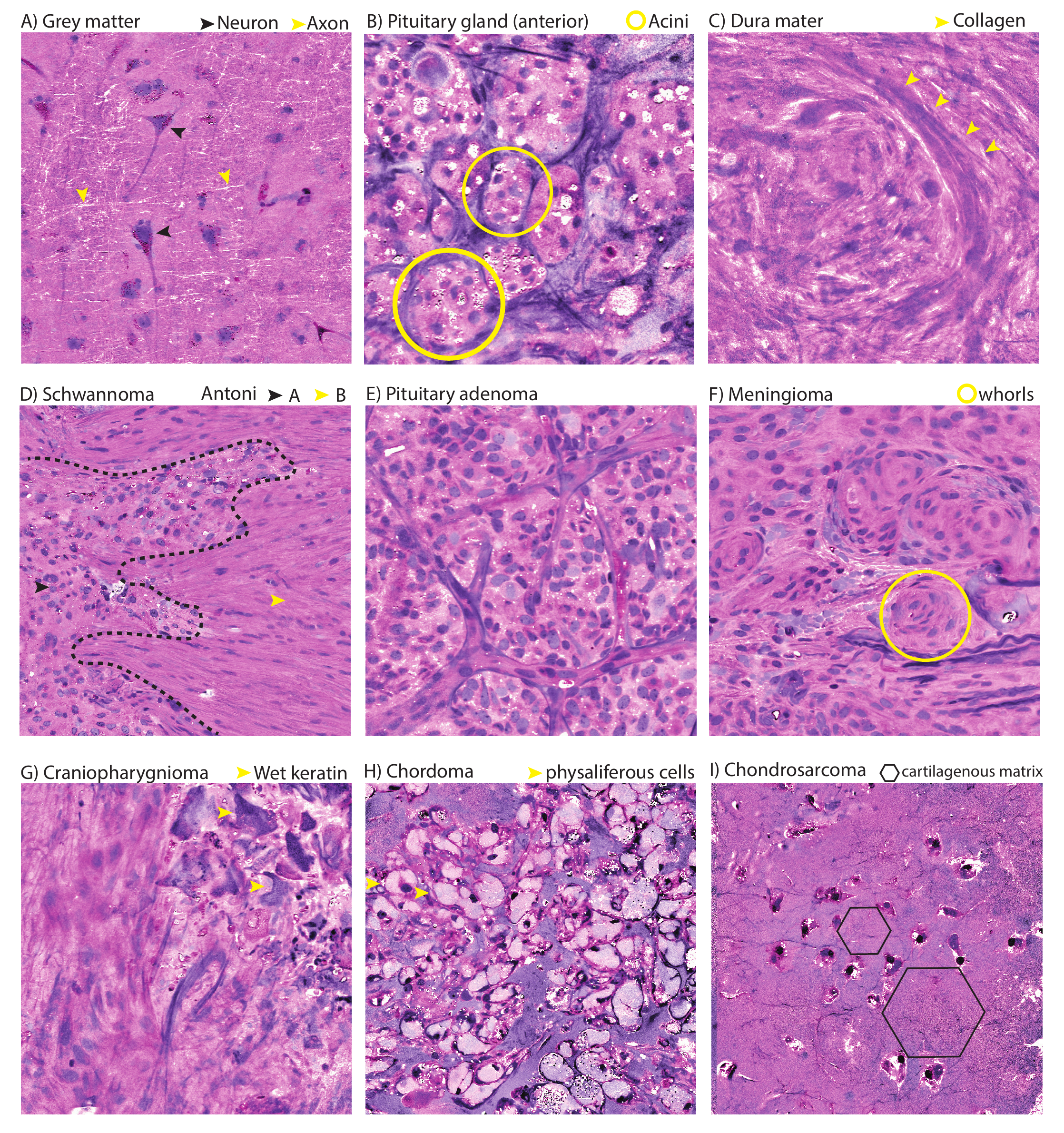}
\caption{
SRH of skull base tumors shows cytologic and histoarchitectural
features. Diagnostic features of normal skull base parenchyma and
skull base tumors are imaged effectively using SRH.
\tbf{A,} Normal grey matter shows pyramidal cell bodies of cortical neurons.
Lipid-rich myelinated axons have high 2845 cm\tsc{-1} signal and appear white in
our virtual H\&E color scheme.
\tbf{B,} Normal anterior pituitary gland has acinar histoarchitecture with
intact reticulin network.
\tbf{C,} Skull base dura is mainly acellular, fibrous tissue with collagenic
and elastic fibers.
\tbf{D,} Schwannoma (vestibular schwannoma shown) shows classic spindle
cell cytology combined with Antoni A and B histoarchitectural patterns.
\tbf{E,} Pituitary adenomas show monotonous cytology with loss of acinar
structure.
\tbf{F,} Meningiomas have large nuclei and whorl patterns throughout the specimen.
\tbf{G,} Adamantinomatous craniopharyngiomas are complex specimens and
uniquely show wet keratin.
\tbf{H,} Clival chordomas have bubbly, physaliferous cells.
\tbf{I,} Chondrosarcomas show chondrocytes embedded in a dense cartilaginous matrix. 
}
\label{fig:panel}
\end{figure*}

\subsection{Stimulated Raman Histology}
All images were obtained using a clinical fiber laser–based stimulated Raman
scattering (SRS) microscope \cite{hollon2020near, freudiger2014stimulated}. The
NIO Laser Imaging System (Invenio Imaging, Inc) is delivered ready to use for
image acquisition and requires a single technician to operate with minimal
training. Viewing SRH images can be performed directly in the operating room or
remotely through medical center radiographic system or cloud-based viewer.
Fresh, unprocessed, surgical specimens are excited with a dual-wavelength fiber
laser as specified in our previous publications \cite{orringer2017rapid,
hollon2020near}. These specifications allow for imaging at Raman shifts in the
range of 2800 to 3130 cm\tsc{-1}. The NIO Imaging System was used to acquire
all images in the testing set \cite{hollon2020near}. For SRH, 2850 and 2950
cm\tsc{-1} are the wavenumbers used to acquire the 2 channel images Lipid-rich
regions (eg, myelinated white matter) demonstrate high SRS signal at 2845
cm\tsc{-1} because of CH2 symmetric stretching in fatty acids. Cellular regions
produce high 2930 cm\tsc{-1} intensity and large signal 2930 to 2845 ratios to
high protein and nucleic acid content. A virtual hematoxylin and eosin (H\&E) color
scheme is applied to transform the raw SRS images into SRH images for clinical
use and pathological review.

SRH combined with AI is an off-label use of the NIO Laser Imaging System. The
AI and algorithms discussed are for research purposes only and have not been
reviewed or approved by the US Food and Drug Administration.

\subsection{Image Dataset and Data Preprocessing}

SRH imaging was completed using 2 imaging systems: a prototype clinical SRH
microscope \cite{orringer2017rapid} and the NIO Imaging System. All collected
clinical specimens were imaged in the operating room using our SRH imagers. In
addition, we used cadaveric specimens of normal tissue (brain, dura, and
pituitary gland) to improve our classifiers ability to detect normal tissue and
avoid false-positive errors. Specimens compromised by hemorrhage, excessive
coagulation, or necrosis were excluded. For image preprocessing, the 2845
cm\tsc{-1} image was subtracted from the 2930 cm image, and the resultant image
was concatenated to generate a 3-channel SRH image (2930 cm\tsc{-1} minus 2845
cm\tsc{-1}, red; 2845 cm\tsc{-1}, green; and 2930 cm\tsc{-1}, blue). A
$300\times 300 \text{pixel}^2$ nonoverlapping sliding-window algorithm was used
to generate image patches. Our laboratory has previously trained a neural
network model that filters images into 3 classes for automated patch-level
annotation: normal brain, tumor tissue, and non-diagnostic tissue
\cite{hollon2020near, hollon2021rapid}. Normal dura was included in the
nondiagnostic class because it lacks cytological features (Figure
\ref{fig:workflow}).

\subsection{Model Training}
Only tumor classes with $>15$ patients were included: pituitary adenomas,
meningiomas, schwannomas, primary central nervous system lymphoma, and
metastases. Normal classes included normal brain (gray matter and white matter)
and normal pituitary gland (anterior gland and posterior gland). Six hundred
patients were included in the training set.

We implemented the ResNet50 CNN architecture with 25.6 million trainable
parameters for our SRH feature extractor \cite{he2016deep}. Three loss
functions were used for model training: supervised categorical cross-entropy,
self-supervised contrastive \cite{chen2020simple}, and supervised contrastive
\cite{khosla2020supervised}. The general contrastive loss function is
\begin{multline}
    \ell_\mathrm{contrastive}(\mbf{z}_x, \mbf{p}_x, \mathcal{N})=\\
-\log\frac{\exp\left(\mathrm{sim}(\mbf{z}_x,\mbf{p}_x)/\tau\right)}
{\sum_{\mbf{n} \in \mathcal{N}}\exp\left(\mathrm{sim}(\mbf{z}_x,\mbf{n})/\tau\right)}
\end{multline}
where $\mbf{z}_x=f(X)$ is the vector representation of image $X$ after a feedforward
pass through the SRH feature extractor, $\mbf{p}_x$ is the representation of positive
examples for image $X$, and $\mathcal{N}$ is the set of negative examples for image $X$
(Figure \ref{fig:workflow}B). Positive examples can be transformations of the
same image (self-supervised) or different images sampled from the same class
(supervised). The feature extraction model produces a 2048-dimension feature
vector for each input image, and each feature vector is further projected down
to 128 dimensions before the cosine similarity metric (sim) is computed.
Contrastive loss functions have some theoretical advantages over cross-entropy
(ie, robustness to label noise), and we hypothesize that contrastive
representation learning is ideally suited for patch-based classification. The
contrastive learning models were optimized using stochastic gradient descent,
and each model was trained using a batch size of 176 for 4 days on 8 Nvidia
GeForce RTX 2080 Ti graphical processing units (GPUs). After the feature
extraction model training was completed, these features were classified using a
linear classifier trained using cross-entropy loss (see Figure
\ref{fig:workflow}C). Each linear classification layer was trained using the
Adam optimizer and a batch size of 64 for 24 hours on 2 Nvidia GeForce GPUs. We
compared our approaches with a conventional model trained using cross-entropy
and a batch size of 64 for 24 hours on 2 Nvidia GeForce GPUs.

\newcommand{\mulcccol}[1]{\multicolumn{3}{c}{#1}}
\begin{table*}[ht]
    \centering
    \begin{tabular}{cccc|ccc|ccc}\hline
        \multirow{2}{*}{} & \mulcccol{Patch} & \mulcccol{Slide} & \mulcccol{Patient} \\\cline{2-10}
                        & Acc   & Top 2 & MCA   & Acc   & Top 2 & MCA   & Acc   & Top 2 & MCA\\\hline
        CE              & 0.830 & 0.930 & 0.822 & 0.871 & 0.951 & 0.899 & 0.915 & 0.958 & 0.931\\
        SSL + Linear    & 0.599 & 0.781 & 0.567 & 0.769 & 0.894 & 0.772 & 0.831 & 0.924 & 0.824\\
        SupCon + Linear & \tbf{0.866} & \tbf{0.953} & \tbf{0.864} & \tbf{0.914} & \tbf{0.969} & \tbf{0.920} & \tbf{0.966} & \tbf{0.983} & \tbf{0.934}\\\hline
    \end{tabular}
    \caption{Model Performances on Held-Out, Multicenter SRH Testing Set.
    The bold entries signifies the best performing model in each metric. Acc,
    accuracy; MCA, mean class accuracy; CE, cross-entropy; SSL, self-supervised
    contrastive learning; SupCon, supervised contrastive learning. Top 2,
    correct class was predicted first or second more probable.}
    \label{tab:results}
\end{table*}

\begin{figure*}[ht]
\centering
\includegraphics[width=\textwidth]{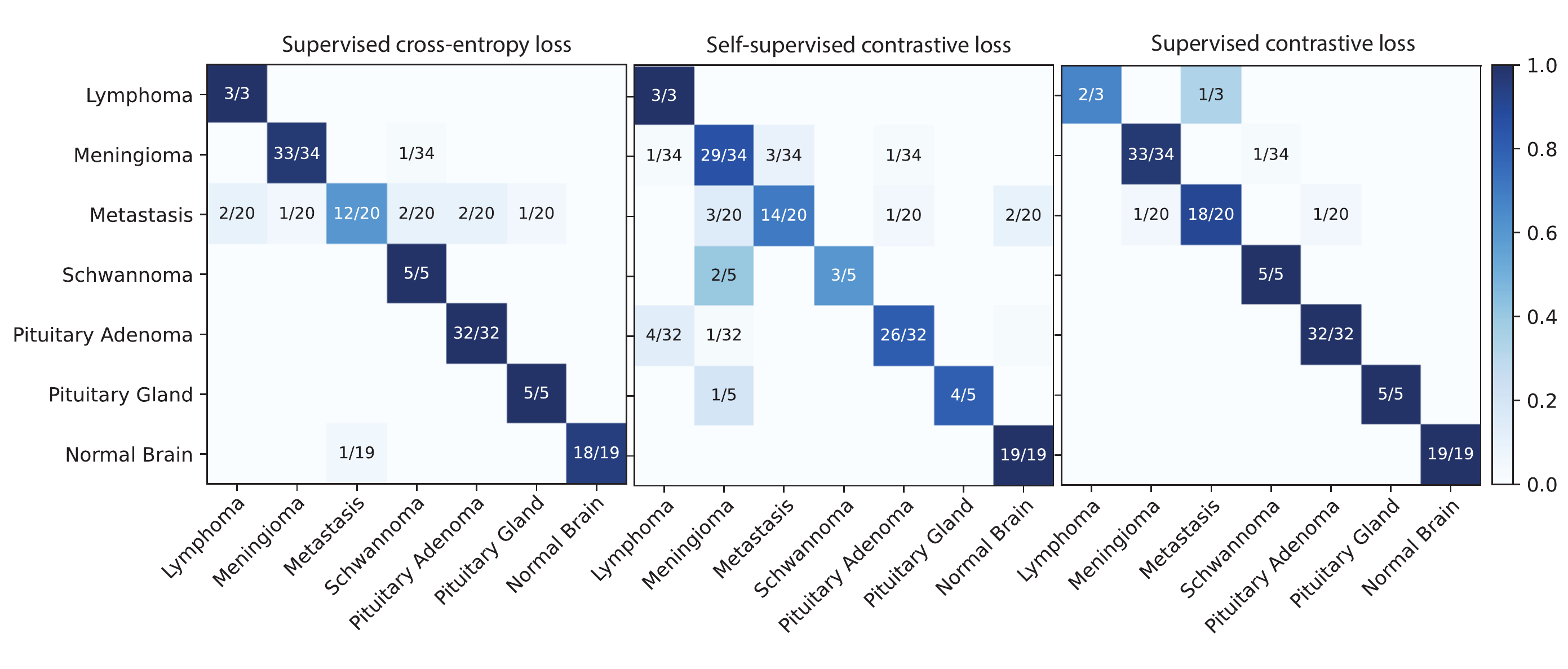}
\caption{Automated intraoperative classification of skull base tumors.
Confusion matrices for each of the 3 training strategies on our held-out,
multicenter, testing set. Supervised cross-entropy achieved an overall
diagnostic accuracy of 91.5\%. Most of the errors occurred in the metastatic
tumors class, with a class accuracy of 60.0\%. Self-supervised contrastive
learning (learning without class labels) performed expectedly worse but still
reached an accuracy of greater than 83\%. Our top-performing model was trained
using supervised contrastive learning, with an overall accuracy of 96.6\% and 2
errors in the metastasis class.
}
\label{fig:cm}
\end{figure*}

\subsection{Model Testing}
We randomly held out 20\% of our data as a testing data set consisting of 118
patients and 489 whole slides. Similar to our training data preparation,
$300\times300$ pixel patches were generated from a whole-slide image, and each
patch underwent a feedforward pass through our trained models to compute a
probability distribution over the output classes. To compute the
whole-slide–level or patient-level accuracy, we summed the patch-level
probability distributions for each whole slide or patient, respectively. The
aggregated probabilities were then re-normalized to compute the final
slide–level or patient-level class probabilities. This ``soft" aggregation of
the classification is superior to ``hard" aggregation of the patches, such as a
simple majority voting procedure, because it takes into account the full
probability distribution for each patch \cite{hollon2020near}.

\subsection{SRH Semantic Segmentation of Skull Base Tumors}
We have previously developed a method for segmenting SRH images
using patch-level predictions \cite{hollon2020near, hollon2021rapid}.
This technique integrates a local neighborhood of overlapping patch prediction
to generate a high-resolution probability heatmap. In a previous study, we
implemented a 3-channel (RGB) probability heatmap which included spatial
information for tumor, normal brain, and nondiagnostic predictions. In this
study, we used a novel technique that generated a 2-channel image with the
predicted tumor class (eg, pituitary adenoma or craniopharyngioma) as the first
channel (ie, red) and the most probable nontumor class (eg, normal pituitary,
normal brain, and nondiagnostic) as the second channel (ie, blue). This method
has an advantage in the setting of skull base tumors by allowing the nontumor
class to vary depending on the surgical specimen. For example, it will
automatically produce a meningioma-normal dura margin heatmap based on the
predicted meningioma diagnosis.

\subsection{Data Availability}
The data that support the findings of this study are available from the
corresponding authors on reasonable request.

\section{Results}
\subsection{Diagnostic Features of Skull Base Tumors}
We first assessed the ability of SRH to effectively capture the diagnostic
features of normal skull base parenchyma and skull base tumors. Figure
\ref{fig:workflow}A shows the general workflow for obtaining SRH images.
Figures \ref{fig:panel}A-\ref{fig:panel}C show the SRH images of normal brain,
anterior pituitary gland, and skull base dura. Classic histological features
are seen, including neuronal cell bodies in gray matter, acinar
histoarchitecture in pituitary gland, and dense collagen extracellular matrix
in dura. Meningiomas, pituitary adenomas, and schwannomas are the most common
skull base tumors encountered (Figure \ref{fig:panel}D-\ref{fig:panel}F). SRH
captures spindle cell cytology and Antoni histoarchitectural patterns in
schwannomas, monotonous hypercellularity in pituitary adenomas, and meningioma
whorls. Less common and malignant tumors are shown in Figure
\ref{fig:panel}G-\ref{fig:panel}I. Wet keratin is well-visualized in
adamantinomatous craniopharyngiomas. Bubble, physaliferous cells are abundant
in clival chordomas. Chondrocytes embedded in a dense cartilaginous matrix are
seen in skull base chondrosarcomas.

\subsection{Automated Classification of Skull Base Tumors Using SRH}
After determining that SRH can effectively capture the diagnostic features in
SRH images, we then trained our CNN using the 3 representation learning methods
(Figure \ref{fig:workflow}B). All models were trained for 4 days and then
tested on our held-out multicenter data set (Table \ref{tab:results}).
We evaluated our model at the patch, slide, and patient levels using overall
top-1 accuracy, top-2 accuracy, and mean class accuracy. Using these metrics,
the model trained using
supervised contrastive representation learning had the best overall
performance, with top scores in all 3 metrics. Our supervised contrastive model
achieves a patient-level diagnostic accuracy of 96.6\% (114 of 118 patients)
and a mean class accuracy of 93.4\%. These results outperformed our
cross-entropy model and significantly improved on our previous results (Figure
\ref{fig:cm}) \cite{hollon2020near}. An important finding was that the
metastatic tumor class was a major source of diagnostic errors for the
cross-entropy model. We believe that this represents the inability of
cross-entropy to effectively represent classes with highly diverse image
features (eg, melanoma vs adenocarcinoma vs squamous cell carcinoma).

\begin{figure*}[ht]
\centering
\includegraphics[width=.8\textwidth]{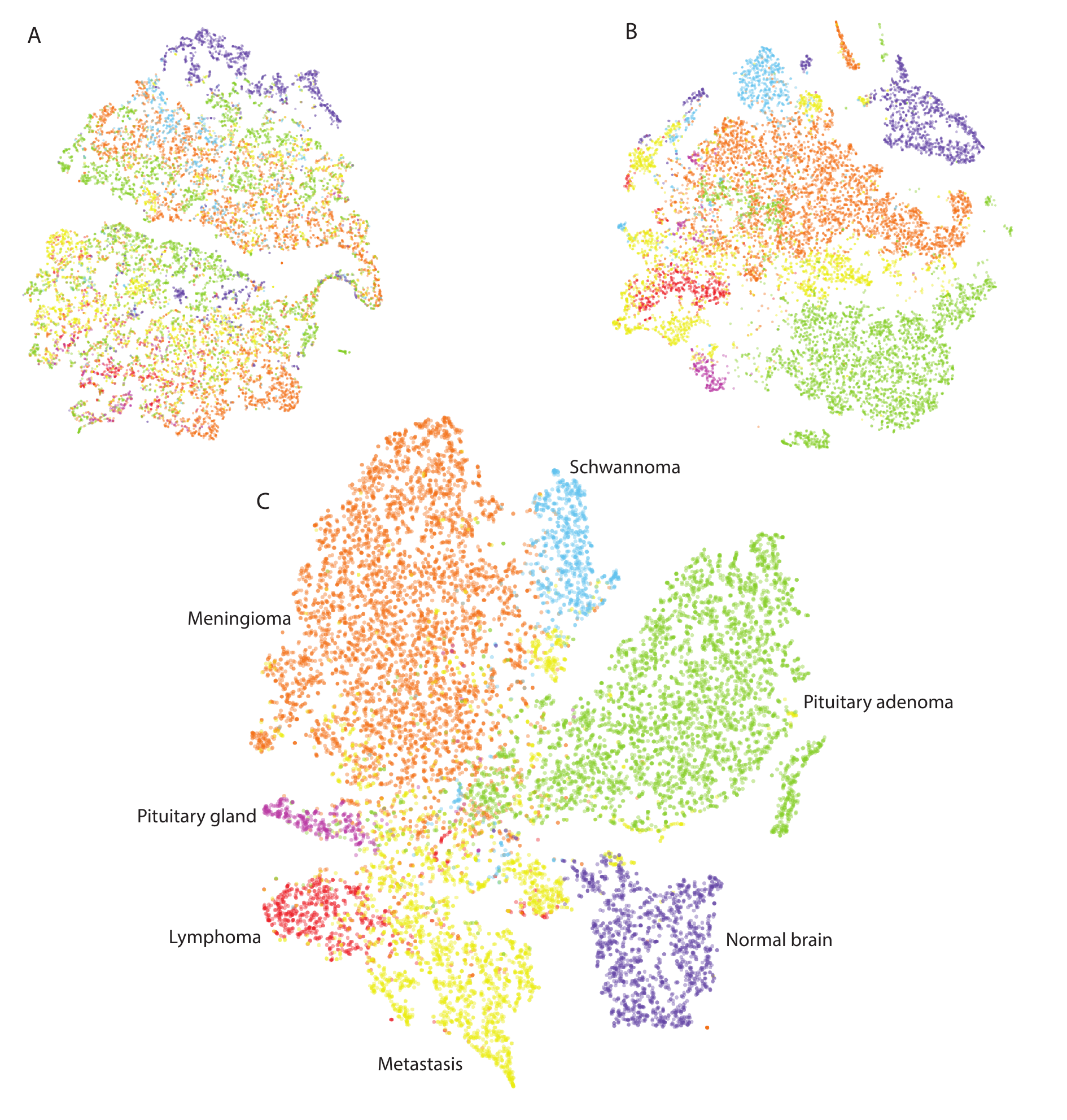}
\caption{Contrastive representation learning t-distributed stochastic neighbor
embedding(tSNE) of classes. This is a tSNE plot of SRH patch representations
from convolutional neural networks trained using \textbf{A,} self-supervised
contrastive, \textbf{B,} cross-entropy, and \textbf{C,} supervised contrastive
loss functions. Each point represents a single SRH patch randomly sampled from
our testing set. Consistent with our diagnostic accuracy results, discrete
class clusters are most discernible in our supervised contrastive
representations, including the metastatic tumor class. Note that the tSNE
algorithm does not depend on class labels, and the color coding is used to
demonstrate that data clusters correspond to tumor classes.}
\label{fig:tsne}
\end{figure*}

\subsection{Visualizing Learned SRH Representations}
We aimed to qualitatively evaluate how effectively the models represented our
SRH images. We used a data visualization technique called t-distributed
stochastic neighbor embedding, which projects high-dimensional data onto a
2-dimensional plane by preserving the local patterns in the data. Data points
with similar representations are located in close proximity, forming discrete
clusters. Compared with cross-entropy or self-supervised contrastive learning,
the supervised contrastive model shows the most well-formed clusters that match
tumor diagnoses (Figure \ref{fig:tsne}). The most salient improvement is how
much more effectively the metastatic class is clustered; contrastive
representation learning explicitly enforces that the model learns image
features which are common to each tumor class, regardless of how diverse the
underlying pathology may be (eg, melanoma vs adenocarcinoma).
 
\begin{figure*}[ht]
\centering
\includegraphics[width=\textwidth]{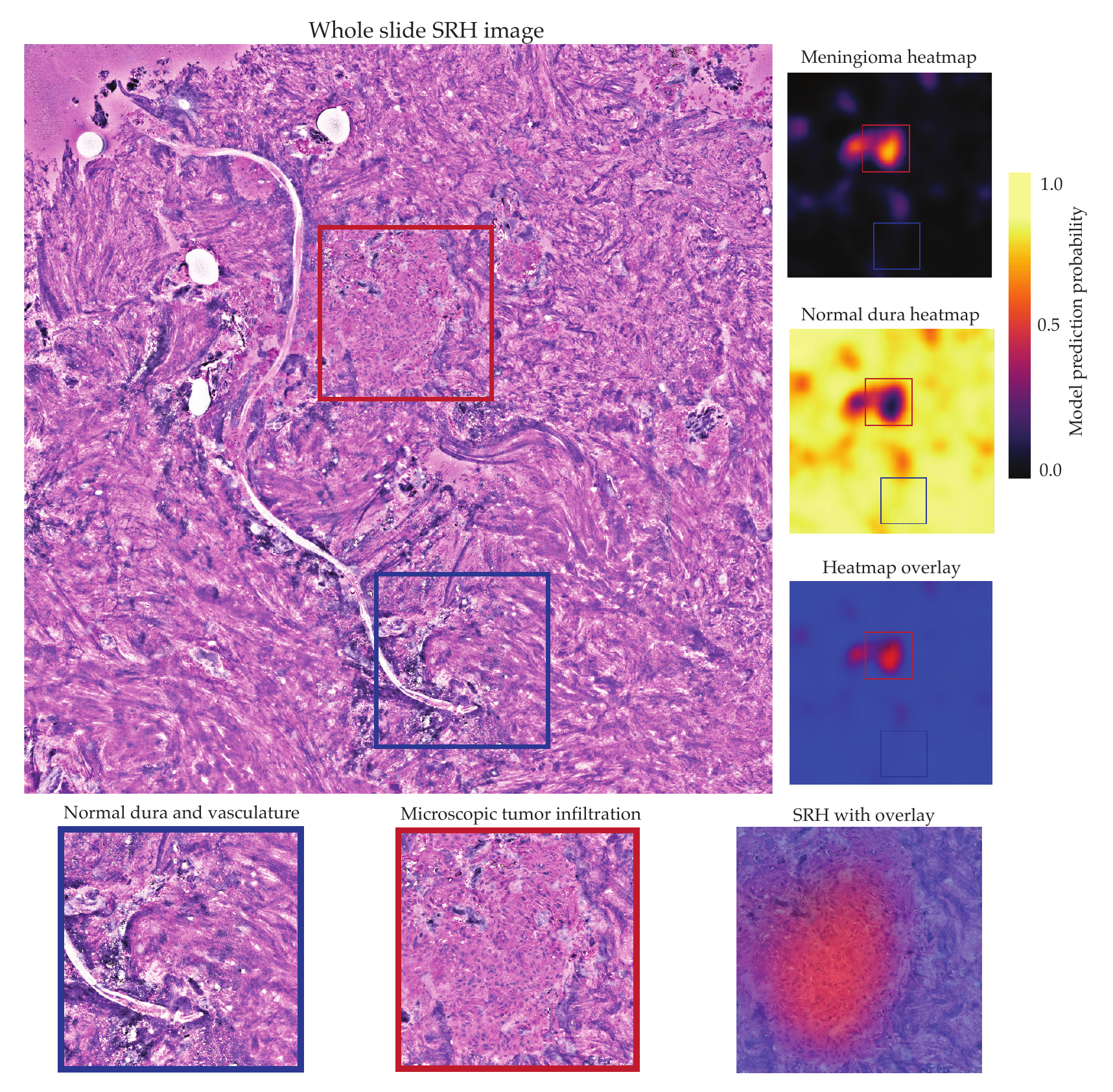}
\caption{Automated detection of microscopic tumor infiltration. Whole-slide SRH
image of grossly normal dura sampled after resection of a tuberculum sellae
meningioma. Microscopic tumor infiltration was detected by our training model,
as shown by the predicted meningioma heatmap over the entire whole-slide image.
Most of the specimen is normal dura with the exception of several small regions
of clear meningioma infiltration. Our predicted heatmaps can be converted into
a colored transparency overlay to be used when reviewing the SRH images
intraoperatively. Heatmaps provide spatial information and serve as an
additional level of decision support for evaluating intraoperative specimens.
}
\label{fig:tumorinfil}
\end{figure*}

\begin{figure*}[ht]
\centering
\includegraphics[width=.8\textwidth]{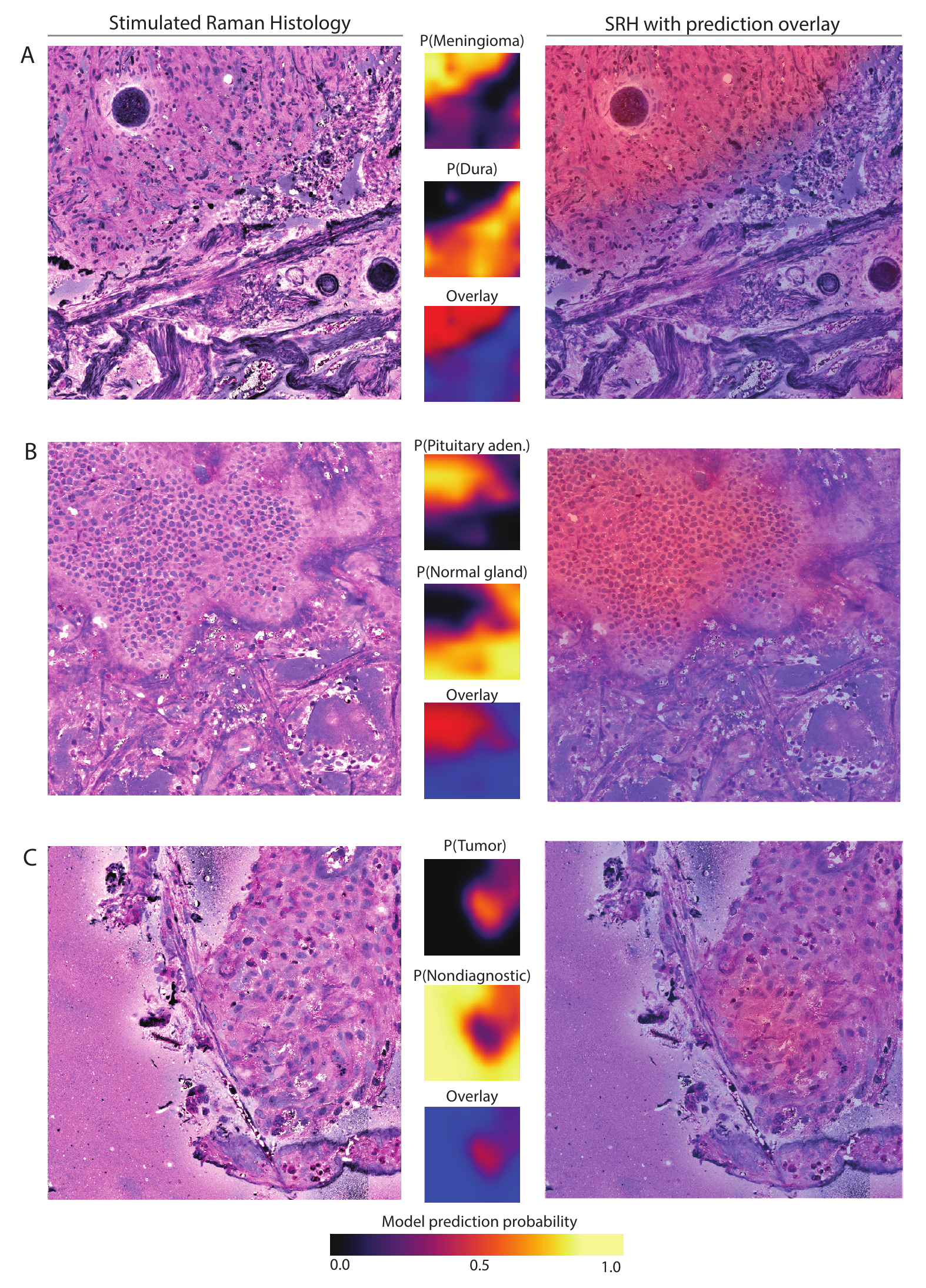}
\caption{SRH semantic segmentation identifies tumor-normal margins and
diagnostic regions. \textbf{A,} SRH image of a meningioma-normal dura margin.
Corresponding prediction heatmaps show excellent delineation of tumor regions
adjacent to normal tissue. \textbf{B,} SRH image of pituitary adenoma-normal
pituitary gland margin. Clear region of monotonous, hypercellular pituitary
adenoma adjacent to normal acinar structure of the anterior pituitary gland.
\textbf{C,} SRH image of a papillary craniopharyngioma. Although our model was
not trained on craniopharyngiomas, it is able to detect regions of tumor and
discriminate them from nondiagnostic, acellular regions. Detection of
diagnostic regions in SRH images can aid in intraoperative interpretation of
large and complex tumor specimens.}
\label{fig:margin}
\end{figure*}

\begin{figure*}[ht]
\centering
\includegraphics[width=.8\textwidth]{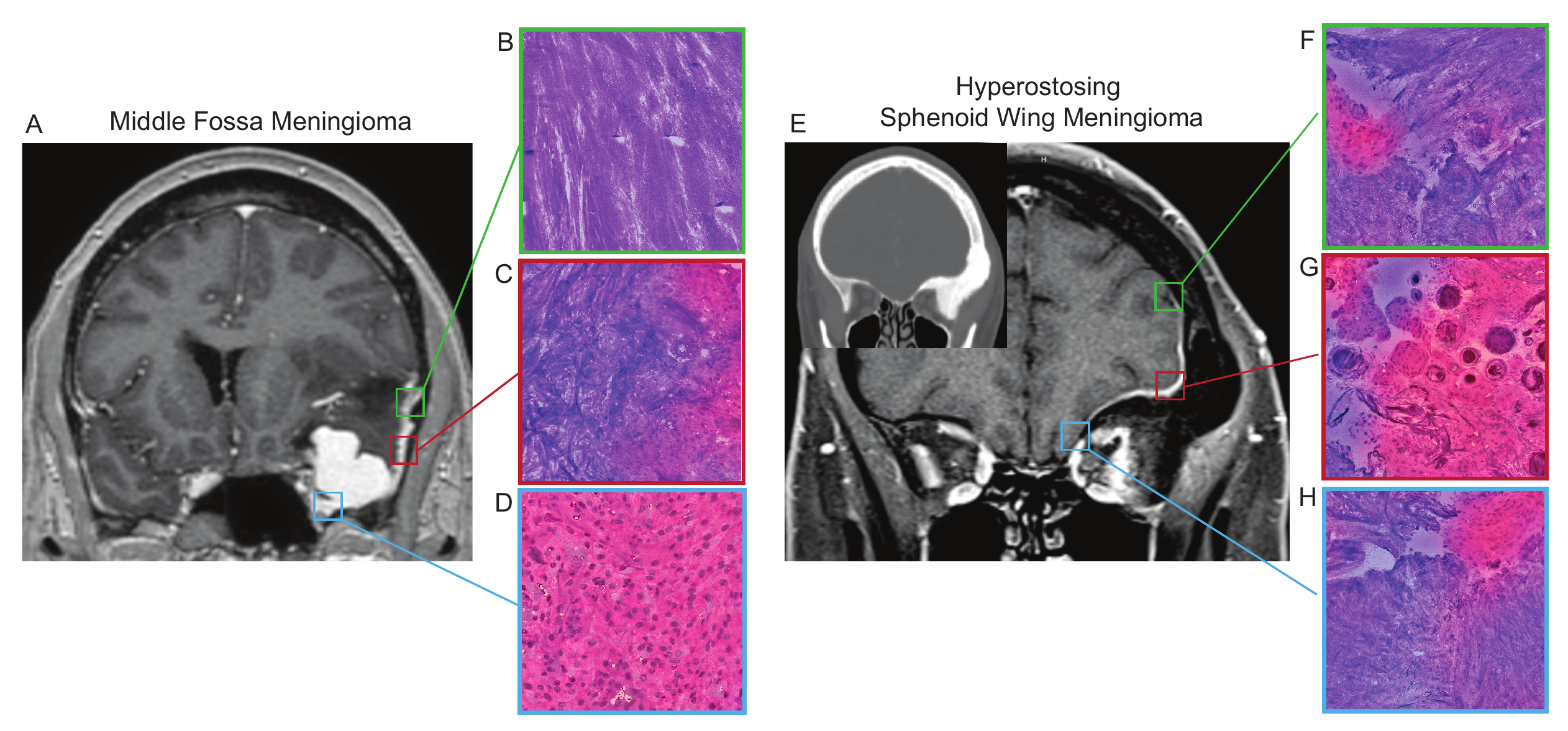}
\caption{Automated analysis of meningioma margins in the clinical setting.
\textbf{A,} A patient with a skull base meningioma arising from the floor of
the middle fossa with an enhancing dural tail extending superiorly along the
temporal lobe. Patient underwent a left pterional craniotomy for tumor
resection. Dural margins were sampled throughout the resection. \textbf{B,} AI
analysis of dura sampled within the dural tail superiorly did not identify
microscopic tumor infiltration. \textbf{C,} Sampled specimen located adjacent
to the dural attachment did show multiple regions of microscopic tumor
infiltration. \textbf{D,} Dense tumor was found infiltrating into the lateral
cavernous sinus and was resectioned. \textbf{E,} A patient with a left en
plaque sphenoid meningioma causing significant hyperostosis and proptosis.
\textbf{F,} Grossly normal dura sampled over the frontal lobe showed dense
regions of microscopic meningioma infiltration. \textbf{G,} Sphenoid wing dura
showed classic psammoma bodies and microcalcification. \textbf{H,} AI analysis
of grossly and radiographically normal dura over the orbital roof detected
microscopic tumor infiltration, and the dura was resected up to the ipsilateral
cribriform plate.
}
\label{fig:mening}
\end{figure*}
\subsection{Detection of Microscopic Tumor Infiltration in Skull Base Specimens}
Using a patch-based classification method allows for a computationally
efficient whole-slide SRH semantic segmentation method. SRH segmentation allows
for improved image interpretation by surgeons and pathologists by providing
spatial information along with the predicted diagnosis (Figure
\ref{fig:tumorinfil}). Moreover, regions of microscopic tumor infiltration can
be automatically detected and highlighted in SRH images. Tumor infiltration can
be identified using the patch-level predictions (Figure \ref{fig:margin})
\cite{hollon2020near, hollon2021rapid}. Importantly, detection of meningioma
infiltration into grossly normal dura can improve extent of resection and
potentially decrease recurrence rates. Our model detected microscopic tumor
infiltration during skull base meningioma surgery (Figure \ref{fig:mening}).
Some dural regions with contrast enhancement (ie, dural tails) did not show
evidence of microscopic tumor infiltration, whereas other dural regions with no
enhancement had clear evidence of meningioma involvement. These results
demonstrate both the feasibility and the importance of microscopic evaluation
of meningioma tumor margins.

\section{Discussion}
In this study, we show that the combination of SRH and AI can provide an
innovative pathway for intraoperative skull base tumor diagnosis and detection
of microscopic tumor infiltration. We were able to achieve a +5.1\% boost in
diagnostic classification accuracy using contrastive representation learning
compared with our previous AI training methods using cross-entropy. The model
effectively identified regions of microscopy brain tumor infiltration and
tumor-normal margins in meningioma SRH images.

Over the previous decade, the applications of AI in clinical medicine and
neurosurgery have grown tremendously. Humanlevel diagnostic accuracy for image
classification tasks has been achieved in multiple medical specialties,
including ophthalmology \cite{gulshan2016development},
radiology\cite{titano2018automated}, dermatology
\cite{esteva2017dermatologist}, and pathology \cite{coudray2018classification,
lu2021ai}. AI for intraoperative diagnostic decision support has been combined
with mass spectrometry \cite{calligaris2015maldi, santagata2014intraoperative},
optical coherence tomography \cite{juarez2019ai}, infrared spectroscopy
\cite{hollon2018shedding, uckermann2018optical}, and Raman spectroscopy
\cite{jermyn2015intraoperative, kast2015identification}. We believe that the
combination of advanced biomedical optical imaging and the latest discoveries
in AI has the potential to provide accurate and realtime decision support for
surgeons and pathologists.

\subsection{Limitations}

A limitation of our study is the limited subset of skull base tumors,
consisting of the most common skull base tumors and the most common
``look-a-like" lesions. We aimed to determine whether, given a sufficient
amount of training data, we could develop an alternative diagnostic system
using SRH and AI. Because additional SRH training data become available for
rare tumors, future studies will include additional skull base tumor diagnoses.
Our proposed contrastive representation learning method is able to accommodate
additional diagnostic classes without changing the training methodology
described here.

\subsection{Future Directions}
Future directions include moving beyond histopathological diagnosis toward
phenotypic and molecular characterization of brain tumors. The proposed model
training technique is flexible, and data labels/model output can be easily
changed or extended to include tumor grade, proliferation indices, and
molecular diagnostic mutations. In addition, access to fresh tumor specimens
provides a unique opportunity to develop optical imaging–based prognostic
biomarkers that have the potential to predict response to treatment (eg,
immunotherapy) and long-term clinical outcomes better than standard diagnostic
methods alone.

\section{Conclusion}
Rapid intraoperative margin delineation in both benign tumors and malignant
skull base tumors, including chordomas and sinonasal carcinomas, has the
potential to improve recurrence-free and overall survival. This study
demonstrated the general feasibility of using SRH and AI for the detection of
microscopic tumor infiltration in real time at the surgical bedside. We applied
these methods specifically to meningiomas because intraoperative Simpson
grading is at risk for underestimating residual tumor, especially for grade I
and II meningiomas \cite{ueberschaer2021simpson}. The proposed method may
reduce residual tumor burden through rapid microscopic assessment of meningioma
specimens.

\section*{Acknowledgements, Funding, and Disclosures}
We would like to acknowledge Tom Cichonski for his editorial contributions.

The study was partially funded by R01CA226527 (Dr Orringer).

Dr Orringer and Dr Hollon are shareholders in Invenio Imaging, Inc. Dr Pandian
is an employee of Invenio Imaging, Inc. Dr Freudiger is an employee, executive,
and shareholder in Invenio Imaging, Inc. The other authors have no personal,
financial, or institutional interest in any of the drugs, materials, or devices
described in this article. Dr Orringer has also received grants/payments from
NX Development Corporation, Stryker Instruments, Designs for Vision, and
DXCover (for serving on the Scientific Advisory Board).

\bibliographystyle{unsrt}
\bibliography{skullbase.bib}

\end{document}